\documentclass[a4paper]{article}

\usepackage[ruled]{algorithm2e}
\usepackage{graphicx}

\SetAlFnt{\small}
\SetAlCapFnt{\small}
\SetAlCapNameFnt{\small}
\SetAlCapHSkip{0pt}
\IncMargin{-\parindent}

\newcommand\blfootnote[1]{%
  \begingroup
  \renewcommand\thefootnote{}\footnote{#1}%
  \addtocounter{footnote}{-1}%
  \endgroup
}

\usepackage{authblk}
\date{}                     %% if you don't need date to appear
\begin{document}
%\graphicspath{ {Figures/} }

% Page heads
\markboth{T.Potok et al.}{A Study of Complex Deep Learning Networks on High Performance, Neuromorphic, and Quantum Computers}

% Title portion
\title{A Study of Complex Deep Learning Networks on High Performance, Neuromorphic, and Quantum Computers}
\author[1]{Thomas E. Potok}
\affil[1]{Oak Ridge National Laboratory}
\author[1]{Catherine Schuman}
% \affil{Oak Ridge National Laboratory}
\author[1]{Steven R. Young}
% \affil{Oak Ridge National Laboratory}
\author[1]{Robert M. Patton}
% \affil{Oak Ridge National Laboratory}
\author[2]{Federico Spedalieri}
\affil[2]{USC Information Sciences Institute}
\author[2]{Jeremy Liu}
% \affil{USC Information Sciences Institute}
\author[2]{Ke-Thia Yao}
% \affil{USC Information Sciences Institute}
\author[3]{Garrett Rose}
\affil[3]{University of Tennessee}
\author[3]{Gangotree Chakma}

\maketitle

\begin{abstract}
Current Deep Learning approaches have been very successful using convolutional neural networks (CNN) trained on large graphical processing units (GPU)-based computers. Three limitations of this approach are: 1) they are based on a simple layered network topology, i.e., highly connected layers, without intra-layer connections; 2) the networks are manually configured to achieve optimal results, and 3) the implementation of neuron model is expensive in both cost and power. 
In this paper, we evaluate deep learning models using three different computing architectures to address these problems: quantum computing to train complex topologies, high performance computing (HPC) to automatically determine network topology, and neuromorphic computing for a low-power hardware implementation. We use the MNIST dataset for our experiment, due to input size limitations of current quantum computers. 
Our results show the feasibility of using the three architectures in tandem to address the above deep learning limitations. We show a quantum computer can find high quality values of intra-layer connections weights, in a tractable time as the complexity of the network increases; a high performance computer can find optimal layer-based topologies; and a neuromorphic computer can represent the complex topology and weights derived from the other architectures in low power memristive hardware. 
\blfootnote{
Notice: This manuscript has been authored by UT-Battelle, LLC under Contract No. DE-AC05-00OR22725 with the U.S. Department of Energy. The United States Government retains and the publisher, by accepting the article for publication, acknowledges that the United States Government retains a non-exclusive, paid-up, irrevocable, world-wide license to publish or reproduce the published form of this manuscript, or allow others to do so, for United States Government purposes. The Department of Energy will provide public access to these results of federally sponsored research in accordance with the DOE Public Access Plan (http://energy.gov/downloads/doe-public-access-plan).}

\end{abstract}

\section{Introduction}

Deep Learning is inspired by the networks of neurons in the visual cortex of the brain. Early versions of these neural networks have been simulated on a computer to analyze imagery. While promising, a significant limitation of this approach has been the computational time required to train or optimally set the weights within a network. Graphical processing units (GPUs) have provided a significant speedup in training, due to their ability to perform multiple simple network weight calculations in parallel, which allows for larger and more complex networks to be studied. These more complex networks containing multiple hidden layers are known as deep learning networks.

In this paper we explore the current limitations of deep learning, and test potential solutions on the emerging computational architectures of high performance computing, quantum computing, and neuromorphic computing. For this study, we will focus on convolutional neural networks (CNNs) and Boltzmann Machines (BMs).

% * <tjtt1987@gmail.com> 2017-01-19T16:20:28.186Z:
% 
% Steven, can you add about 1/2 page? Give some general background on CNNs
% DONE
% ^ <tjtt1987@gmail.com> 2017-02-23T21:33:02.164Z.
There are multiple designs for deep learning networks, with, CNNs being the most widely used deep learning models \cite{lecun1998gradient} and are most commonly used for image classification with remarkably good results.
CNNs have many similarities to traditional multi-layer perceptron (MLP) neural networks.
They utilize stochastic gradient descent and backpropagation for training a set of learnable weights in a supervised manner.
The definings feature of convolutional networks are their convolutional layers and pooling layers.
Convolutional layers consist of multiple sets of weights, or kernels, that are convolved with their input to form a set of feature maps.
Pooling layers are used to subsample the outputs of a convolutional layer to produce a smaller feature map, usually by using an max or average operation.
These layers combine to utilize shared weights and pooling to take advantage of the locality that exists within images.
The concept of shared weights builds upon the assumption that if a feature is useful in one location of an image it is also useful in other locations. 
Thus, the network need not calculate a unique set of weights for every position in the image. 
Since these weights are shared, activations from one position in an image are calculated using the same weights as other locations, and these weights can be pooled in a meaningful way. 
This pooling operation allows the network to be resilient to shifts of features within the image \cite{scherer2010evaluation}.
Subsequent improvements to CNNS, such as dropout \cite{hinton2012improving}, rectified linear units \cite{jarrett2009best,nair2010rectified}, and efficient GPU DL codes \cite{tensorflow2015-whitepaper,theano-full,jia2014caffe}  have allowed CNNs to achieve impressive results on a variety of benchmarks.
These include the ImageNet \cite{russakovsky2015imagenet} dataset where CNNs have surpassed human level performance at object recognition \cite{he2015delving} and the Labeled Faces in the Wild (LFW) \cite{huang2008labeled} dataset where CNNs have surpassed human level performance at face recognition    \cite{schroff2015facenet}.

% * <tjtt1987@gmail.com> 2017-01-19T16:20:48.257Z:
% 
% Steven, need to expand to about 1/2 page, general background on BM
% 
% ^ <tjtt1987@gmail.com> 2017-02-23T21:33:21.694Z.
A Boltzmann Machine (BM) is a recurrent neural network consisting of neurons that make binary stochastic decisions based on the states of their symmetrically connected neuron neighbors \cite{ackley1985learning}. 
Boltzmann Machines are well-suited for solving constraint satisfaction tasks with many weak constraints. 
These tasks include digit recognition, object recognition, compression/coding and natural language processing. 
A BM training algorithm was proposed  in \cite{ackley1985learning}. 
This training algorithm relies on iterations of updating the states until a thermal equilibrium is reached, and then updating weights based on a simple learning rule.
Though this process can be improved by utilizing simulated annealing to reach thermal equilibrium, it is a very slow process for large networks.
As there are $2^N$ possible states the network could take for a network comprised of $N$ neurons along with $2^N$ weights to be learned. Identifying the state of thermal equilibrium and calculating equilibrium statistics becomes a challenge for large networks even for the best computational resources available as the computation grows exponentially with $N$.
As such, the training of a BM is impractical for complex network topologies.  
This has given rise to the development of a Restricted Boltzmann Machine (RBM). 
The RBM network topology is restricted to a bipartite graph \cite{Hinton504}.  
Deep belief networks (DBNs) can be created by composing many RBM layers \cite{Hinton:2006:FLA:1161603.1161605}.

There are currently three main challenges in Deep Learning. 

The first is how to train models with complex topologies that are closer representations of nature. Current deep learning networks limit the  scale and complexity of the neuron models they use. While very successful in solving challenging classification problems, these neuron models are not comparable to those produced by nature. Early deep learning models proposed networks
 which contained intra-layer connections, but proved to be intractable to train on conventional computer systems. We believe that quantum computing may offer a potential solution with the ability to sample from complex probability distributions like those generated by neural networks that contain intra-layer connections.

The second challenge is how to automatically configure a network to an optimal or near optimal topology.  Current deep learning models are created, trained and tested on reference datasets, with high performing network configurations reported in the literature. A significant challenge is how to construct a high performing network from previously unexamined data. GPU-based high performance computing provides an opportunity to train, test, and evolve thousands of deep learning networks to find well performing network hyperparamaters. 

The last challenge is how to natively implement a complex deep learning model using a simulated neuron and synapse hardware architecture, as opposed to a CPU/Memory based model. While conventional computer can and are being used to deploy a deep learning network, the power requirements are high, especially when compared with nature. Neuromorphic devices provide neuron and synapse hardware architecture that incorporates a time component (spike), and when implemented with memristive technology has the potential to run deep learning networks with very low power consumption.

Our hypothesis is that these three deep learning challenges can be addressed through a combinations of quantum, high performance, and neuromorphic computing. To test this hypothesis we use a simple deep learning problem, MNIST,  using a native deep learning network representation for each of the the three computing platforms, i.e., a BM for quantum, a convolutional neural network for  high performance computers, and a spiking neural network for neuromorphic.

This paper will provide a brief background on the challenges of deep learning related to quantum, high performance, and neuromorphic computing, followed by our experimental approach, results and future research.

\section{Related Work}

First we look at the current state of art of quantum, high performance, and neuromorphic computing as related to the challenges in deep learning as stated above. 

\subsection{Quantum Computing}
\label{sec:relatedWork:quantum}
Computing using quantum computers was first discussed by Feynman \cite{Feynman-82} who was motivated by the fact that simulating a quantum system using a classical computer seems to be intractable. Interest in quantum computing increased dramatically with the discovery of the Shor's polynomial quantum algorithm for factoring numbers \cite{Shor-97}, because all known  classical probabilistic factoring algorithms require exponential time. Several approaches to quantum computing were since developed, and they include the well-known quantum circuit model (used by Shor's algorithm), the measurement-based quantum computing model, and the adiabatic quantum computing model \cite{Farhi2000}. All three have been shown to have the same computational power. In this paper, we focus on a restricted form of the adiabatic quantum computing model which performs adiabatic quantum optimization (AQO) to find the minimum energy state of an Ising Hamiltonian system. Actual implementations of adiabatic quantum machines, such as the D-Wave, operate at a finite temperature \cite{johnson2011a}. 

The output of these machines is a sequence of samples from a probability distribution defined by the Ising Hamiltonian system. The ability to draw samples from complex probability distributions is at the core of probabilistic deep learning approaches, like the BM. As stated above, the training of a BM is impractical on traditional computer systems, thus the development of a RBM restricts the network topology to that of a bipartite graphs \cite{Hinton504}. Bipartite graphs allow for techniques like contrastive divergence to efficiently draw approximate samples in linear time from the probability distribution defined by the BM. This potentially enables the evaluation of a network of restricted BM. Sampling is a fundamental building block and part of the inner loop of the Boltzmann learning algorithm. Without the bipartite restriction, sampling a BM with a general topology is a NP-hard problem. Adiabatic quantum machines have the potential to efficiently sample a richer set of graph topologies that are supersets of bipartite graphs. Several approaches to exploit this feature have been attempted using the D-Wave processor for different choices of graph size and topology \cite{adachi2015a,PhysRevA.94.022308,benedetti2017a}.

\subsection{High Performance Computing}
% * <tjtt1987@gmail.com> 2016-07-13T20:59:42.545Z:
%
% Steven - I need a background section on work done in deep learning using complex topologies, and why they do not work well  
% Doing this in the introduction section where we talk about BMs since it doesn't seem to fit well here.
% ^ <tjtt1987@gmail.com> 2016-08-09T18:59:58.026Z.

% Deep learning and HPC
Deep learning, being an early adopter of GPU technology, has benefited greatly from the speedup offered by these accelerated computing devices and has received great support from device manufacturers in the form of deep learning-specific GPU libraries. 
General purpose GPUs are the basic building blocks of today's HPC platforms and next generation machines will rely on them to an even greater degree. 
Thus, deep learning provides a great opportunity to fully utilize these machines, as they will have multiple GPUs per compute node. 
This leaves the question of how to best utilize thousands of GPUs for deep learning, as previous work has only utilized a maximum of 64 GPUs before encountering scaling problems when trying to exploit model parallelism to spread the weights of the network across multiple GPUs \cite{coates2013}. 
HPC provides the unique opportunity to address the problem of network specification. This refers to the problem of deciding upon the set of hyper-parameters needed to specify the network and training procedure in order to apply deep learning to a new dataset. 

For convolutional neural networks, this could involve specifying parameters such as the number of layers, the number of hidden units, or the kernel size. For more general networks, such as RBMs, this could involve defining much more complicated connectivity between neurons.

Previously, it has been shown that HPC can be utilized to optimize the hyper-parameters of a deep learning network \cite{young2015}. 
This work utilized an evolutionary algorithm distributed across the nodes of Oak Ridge National Laboratory's (ORNL's) Titan supercomputer in order to optimize the performance of deep learning algorithms.
Hyper-parameters in deep learning refer to the model parameters, i.e., the activation function used, the number of hidden units in a layer, the kernel size of a convolutional layer, and the learning rate of the solver. As the size of the network grows, the hyper-parameter space grows increasingly larger.
The size of deep learning networks used today have resulted in a hyper-parameter space that cannot be searched on a single machine or a small cluster.
This is a result of the computational complexity of training and evaluating these networks.
Without utilizing the computational capabilities provided by supercomputers, evaluating a sufficient number of hyper-parameter sets to search the enormous hyper-parameter space of these methods would be impossible.

\subsection{Neuromorphic Computing}
\label{sec:relatedWork:neuromorphic}
Neuromorphic computing architectures have historically been developed with one of two goals in mind: either developing custom hardware devices to accurately simulate biological neural systems with the goal of studying biological brains or building computationally useful architectures that are inspired by the operation of biological brains and have some of their characteristics.  In developing neuromorphic computing devices for computational purposes, there have been two main approaches: building devices based on spiking neural networks (SNNs), such as IBM's TrueNorth \cite{cassidy2013_truenorth} or Darwin \cite{shen2016_darwin}, and building devices based on convolutional neural networks, such as Google's Tensor Processing Unit \cite{google_tpu} or Nervana's Nervana Engine \cite{nervana_engine}, to serve as deep learning accelerators.  
The neuromorphic devices that have been built based on SNNs or built to simulate biologically-accurate systems have vastly different characteristics than those that have been built based on deep learning networks, such as CNNs or RBMs. The neurons in SNN-based systems are typically not organized in layers, and there are fewer restrictions on connectivity between neurons.  The neuron and synapse models also differ from those in convolutional neural networks. In SNN-based neuromorphic systems, the neuron is typically some form of spiking neuron, such as a leaky-integrate-and-fire neuron, and the synapses have a delay value in addition to a weight value, thus introducing a temporal component to the processing of the network. 

The primary computational issue associated with SNN-based systems is that very few algorithms that train native networks for those systems have been developed.  Two of the key reasons why algorithms have not been developed are the computational difficulty introduced by broader connectivity and the computational difficulty introduced by  the inclusion of the temporal component in both the neurons and synapses. In fact, one approach for training networks for neuromorphic computers has been to train a CNN offline and then create a mapping process from the CNN to the associated SNN-based neuromorphic hardware \cite{esser2015backpropagation}.

One of the key properties of neuromorphic systems is their potential for more energy-efficient computation.  To achieve energy-efficiency, we (and many others) have explored an implementation of a spiking neural network system utilizing memristors.  Memristors are one of the four fundamental circuit elements. They are ``memory resistors'' in that their resistance can be altered depending on the magnitude of the voltage applied. Likewise, when no voltage is applied across a memristor, the most recent resistance value is retained \cite{williams2008memristor}.  Memristors have similar behavior to biological synapses, and as such, have been frequently utilized to implement neuromorphic systems \cite{jo2010nanoscale,kim2011functional,prezioso2015training}.

\section{Approach}

The three platforms we are studying, quantum, high performance, and neuromorphic computing, are quite different in the way that they process data. Selecting a deep learning problem that can be used on all three is constrained by the amount of data that each can support.  Currently D-Wave supports 1000 qubits, which limits the size of a problem to the inputs and deep learning network. MNIST is a collection of hand-written digits that has been very widely studied in the deep learning community \cite{mnist}. The images of the digits are very small (28 X 28 pixels totaling 784 pixels) that can be analyzed using 1000 qubit of quantum computer as well as the other architectures. 

The next challenge is selecting the type of deep learning network that can be natively supported on the three platforms. While many deep learning methods have been proposed over the years, CNN have consistently provided the highest accuracy on standard datasets, and typically would be the network of choice for such a comparison. However, the quantum architecture provides a native BM representation that does not restrict intra-layer connections, which is computationally impractical for conventional computers. Likewise, for a neuromorphic computer, a spiking neural network provides a native time-based analysis model. Both a strongly connected BM, and time dependent spiking neural network operate quite differently than a CNN, but in conjugation with their respective platforms, provide a distinctive capabilities that we believe can augment or strengthen a CNN model. 

The last challenge is on how to compare the experimental results of the three platforms. A nominal performance comparison of the three approach provides little insight to the deep learning challenges we are addressing. Likewise an accuracy comparison on the all but solved MNIST problem is not helpful either. Instead, we look to compare the capabilities the three architectures provide in addressing the deep learning challenges we stated above.  We believe that a combination of the architectures may provide a significant benefit over that of a single platform.

\begin{figure}[htpb]
  \centering
  \includegraphics[width=0.5\textwidth]{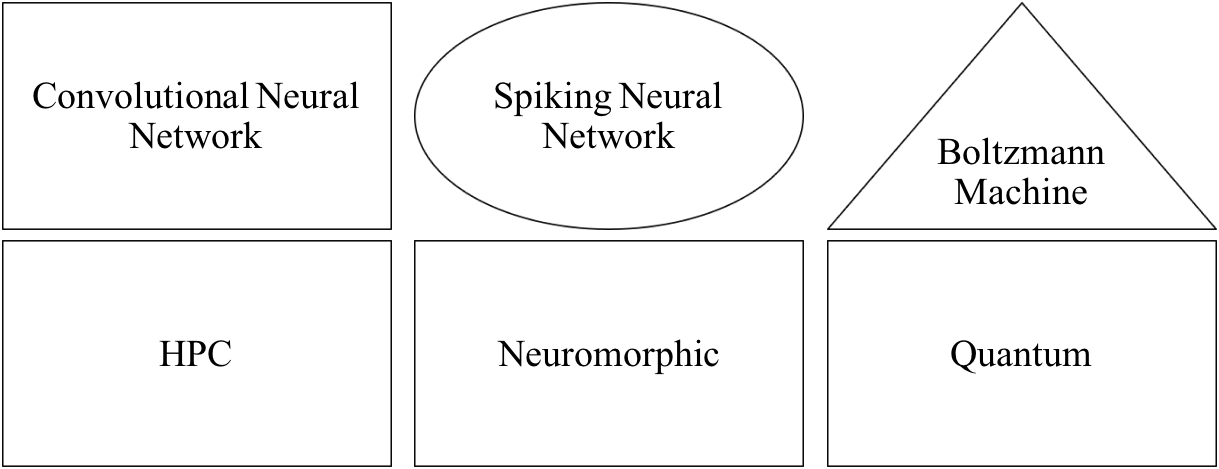}
  \caption{The native deep learning methods for each hardware platform.}
  \label{archFigure}
  \centering
\end{figure}

Figure \ref{archFigure} shows a notional diagram of the deep learning networks that will be applied to each of the different architectures.

Specifically, a quantum computer will used to address the first deep learning challenge we list, namely, complex topologies that are closer representations of nature. Quantum computers have the ability to sample a probability distribution of a network of BM, and may provide a feasible approach to training a highly connected BM network. 
We will use high performance computers to address the second challenge of how to automatically configure a network to an optimal or near optimal topology using evolutionary algorithm to evolve a high performing network. And lastly, we will address the how to natively implement such a model using a neuron and synapse hardware architecture simulation of a memristive based spiking neural network.

\subsection{Quantum}

The quantum computer we are using is a D-Wave adiabatic quantum computer located at the University of Southern California Lockheed Martin Quantum Computing Center. 
We propose a network of BMs to represent the MNIST problem. As discussed in Section \ref{sec:relatedWork:quantum}, a deep learning network of BM has been previously proposed (Deep Boltzmann Machine); however, learning is intractable for a BM with a fully connected topology, since computing expected values over the model requires computing sums over an exponentially large state space. RBMs were introduced to circumvent this issue by discarding couplings between nodes within the same layer. Removing intra-layer couplings introduces conditional independence between nodes within the same layer. This computational advantage comes at the cost of lower representational power. The D-Wave device provides an opportunity to test this approach. First, we will create and train a RBM on D-Wave, applying it to the MNIST handwritten digit classification problem \cite{mnist} to establish a reference result.

Next, we will consider a more complex topology that allows for intra-layer connections between nodes. We call this semi-restricted BM a "Limited Boltzmann Machine" (LBM). The LBM has two layers: one visible and one hidden. With a 1000 qubit, the visible layer requires 784 qubits, leaving a little over 200 qubits for the hidden layer. The visible layer is the same as in the RBM, with no intra-layer couplings. All the nodes of the visible layer are connected to all the nodes of the hidden layer. However, in contrast to the RBM, the LBM's hidden layer is allowed to have intra-layer couplings. 

\begin{figure}[htbp]
  \centering
  \includegraphics[width=0.25\textwidth]{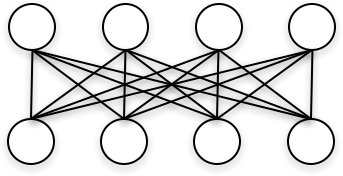}
  \caption{Chimera graphs are composed of 8-qubit cells, with bipartite connectivity.}
  \label{fig:chimeraGraph}
\end{figure}

The hidden layer topology of the LBM is based on the Chimera graph, which represents the underlying connection topology of the D-Wave processor. Chimera graphs are composed of 8-qubit cells. Within each cell, the nodes have a four-by-four fully bipartite connectivity. The bipartite cells are arranged in a grid pattern. The four nodes from one side of the bipartite graph are linked horizontally to two other nodes of adjacent cells in horizontal direction. The other four nodes are linked vertically to two other nodes of adjacent cells in the vertical direction (Figure \ref{fig:chimeraGraph}).

As a consequence, the probability distribution of the hidden layer nodes no longer factorizes when the values of the visible layer nodes are fixed. To estimate the expected values required for the learning process, we used the D-Wave processor to generate samples of the hidden layer configuration and estimate probabilities.

We also included 10 ``classification" digits used to calculate gradients and determine the LBM's output label, i.e., what digit has been read.

We chose 6,000 images from the MNIST database for training the LBM. We use a subset of randomly selected images instead of using the full MNIST dataset due to time considerations. Our D-Wave system is time-shared between different organizations, and within each time-share there are many users. Limiting ourselves to 10\% of the data allowed us to run experiments in a timely manner.

% * <tjtt1987@gmail.com> 2017-01-23T17:44:16.434Z:
% 
% Jeremy - why 6000, how did you choose them, can you describe which 6000 you chose so someone can reproduce the experiment.
% 
% ^ <tjtt1987@gmail.com> 2017-02-27T13:48:03.944Z.

Each 28x28 image is represented by a vector of 784 length, each unit with holding a value representing pixel intensity in the range [0, 1]. In addition to these 784 units representing pixel intensities, we include the previously mentioned 10 additional units to represent the image's digit label (0 through 9) using 1-hot encoding. For example, an image of the digit ``3'' will have label unit 3 marked ``on'' while all other classification nodes will be off). Each image is thus represented with a vector of length 794. When we show each image to the RBM, we hide the image labels and have the RBM attempt to reconstruct the label units. We choose the unit label of highest "on" probability as the RBM's digit classification choice for each image. 

Training weights for the LBM are randomly initialized, just as they are in the RBM. Also as in the RBM, their values are sampled from a standard normal distribution and are updated using the same gradient descent procedure.
% * <tjtt1987@gmail.com> 2017-01-30T18:21:37.912Z:
% 
% Jeremy,
% 
% Can you give a more detailed explaination for  how your read in the images, and how the training weights we set
% 
% ^ <tjtt1987@gmail.com> 2017-03-01T20:55:26.193Z.

\subsection{HPC}
The HPC computer we are using is the ORNL's Titan computer with roughly 300,000 cores, and 18,000 GPUs. This is currently the fastest open science computer in the world. 

Clearly a supercomputer is not needed to solve the MNIST problem; however, a supercomputer is needed to automatically find an optimal deep learning topology for such a problem. Rather than using a trial-and-error method for finding a well performing network topology, we propose to use evolutionary optimization on Titan to evaluate tens of thousands of topologies; therefore, systematically finding the best performing networks on this problem. If achievable, this would solve one of the major challenges in building deep learning networks.

For this experiment we use a CNN as our deep learning network since CNNs are currently producing the top results. We approach the network topology problem of selecting optimal hyper-parameters as a massive search problem, where Titan can be used to quickly search the space.
% * <tjtt1987@gmail.com> 2017-01-23T21:51:01.812Z:
% 
% Robert, Can you say more about the type of evolutionary algorithm you are using? 
% 
% ^ <tjtt1987@gmail.com> 2017-03-01T20:55:31.717Z.

We represent each individual within the population of the evolutionary algorithm (EA) as a single deep neural network or CNN. An individual consists of a genome where the genes represent the various hyper-parameters that define the network topology, i.e., the number of layers, type of layers (convolution, pooling, etc.), and order of the layers. We then apply parameters defined in the genes of the individual to construct and train a deep learning network on the MNIST dataset. The results of the network's performance in testing are then used as the "fitness" of the individual in the EA population, i.e., individual networks that have high accuracy are considered to be the most fit. Typically, generating the results for a single network on a small dataset like MNST will require a modest amount of GPU/CPU time, and memory. However, creating, training, and evaluating tens of thousands networks requires a significant number of GPUs, like those in the Titan high performance computer.

After all the individuals in the population are evaluated, the top performing individuals are selected to generate a new population of individuals that represent the next generation of the EA.
These new generations contain a mix of the well performing hyperparameters from the best performing networks in the population. Successive generations of individuals gradually leads to an improved set of hyperparameters over time.  This method is called Multi-node Evolutionary Neural Networks for Deep Learning (MENNDL).

For this experiment we are looking to automatically discover hyperparameters of a well performing deep learning network on the MNIST dataset. We used a simple EA that limits the search to the number of neurons per layer and the kernel size of convolutional layers.

The network architecture utilized was LeNet \cite{lecun1998gradient}.
This network utilizes 2 convolutional layers, 2 pooling layers, and one hidden fully-connected layer.
This is the network that is most often used with the MNIST dataset in the literature.
% * <tjtt1987@gmail.com> 2017-01-23T22:06:01.831Z:
% 
% Steven, can you briefly summarize the network in a sentence or two?
%  DONE
% ^ <tjtt1987@gmail.com> 2017-03-01T20:58:52.194Z.

We have shown that even with this widely studied MNIST dataset, better hyper-parameters can be found than those widely reported in the literature.  An EA that can evolve the topology provides the opportunity for improved results, and the ability to process more challenging datasets. Such an EA will also provide the opportunity to meaningfully utilize the entirety of Titan's capacity. It will provide challenging data management problems on a machine designed primarily for modeling and simulation, as opposed to these deep learning algorithms which require heavy amounts of data input in addition to heavy computation.

\subsection{Neuromorphic}

A neuromorphic approach to the MNIST problem is not the ideal solution since there is not a temporal component in the task of recognizing a handwritten digit.
To add a temporal component, we use a streaming scan of the digits as input to the SNN. The SNN learns to recognize digits based on this scan pattern.  The goal is to understand the deployment benefits of using an SSN in memristive hardware, as opposed to classification accuracy on this problem. 

We then propose to evaluate the performance of this network on memristor hardware having the potential to enable a low power hardware implementation of a deep learning network, further, with the ability to represent spatial and temporal data. 

As noted in Section \ref{sec:relatedWork:neuromorphic}, there are not very many SNN training methods or training methods that can be applied to neuromorphic networks.  To train both SNN models and neuromorphic networks, we utilize an EA approach to determine the structure (e.g., number of neurons and synapses and how they are connected) and parameters (e.g., weight values of synapses and threshold values of neurons). 

The neuromorphic system we will use to explore the MNIST problem is a memristive implementation of the neuroscience-inspired dynamic architectures (NIDA) system \cite{SCHUMAN201489}. NIDA is a simple SNN model composed of integrate-and-fire neurons and synapses with delays and weights that are affected by processes similar to long-term potentiation and long-term depression in biological brains.  A digital hardware implementation based on NIDA, called Dynamic Adaptive Neural Network Array (DANNA), has also been created and is currently implemented on FPGA with a digital VLSI implementation in progress \cite{schuman2016bica}.  NIDA synapses have analog weight values, while DANNA is restricted to a finite set of digital weight values.  DANNA also has restricted connectivity between neurons, whereas the NIDA model allows for up to fully connected networks. The NIDA model allows for us to study neuromorphic models in software and determine how restrictions different restrictions in hardware (such as weight resolution or connectivity) affect performance. 
% * <tjtt1987@gmail.com> 2017-01-30T18:32:11.470Z:
% 
% Katie, can you give a 2 to 3 sentence description of NIDA and for DANNA
% 
% ^ <tjtt1987@gmail.com> 2017-02-23T21:39:09.206Z.
The EA approach for training networks for the MNIST problem was previously applied to the NIDA SNN \cite{SCHUMAN201489} and to DANNA \cite{schuman2016bica}. For both NIDA and DANNA, an ensemble approach is utilized, where each network in the ensemble is responsible for recognizing a particular digit type.  For example, a network may be trained to recognize zeros, in which case the network will take the handwritten digit image as input and its output corresponds to either ``yes, it is a zero'' or ``no, it is not a zero.''  Using this approach, ensembles that achieve around 90 percent accuracy for NIDA and around 80 percent accuracy for DANNA were created.  

For this work, we extend the prior work by simulating a SNN (specifically, a NIDA network) implemented in memristive hardware in order to demonstrate the potential of significant power reductions for simulating the behavior of neural networks.

\section{Results}

\subsection{Quantum}
% Results of RBM using D-wave

We utilize common parameters to control the learning progress, the same ones found in training RBMs \cite{hinton2010practical}. We chose to conduct training over 25 epochs (one epoch is a complete run over all the training data) instead of 10 epochs to get a better idea of what performance we can potentially achieve. Another parameter is the learning rate, or how much the LBM learns from each example. Setting the parameter too high can cause unstable behavior. This can be thought of as the LBM compensating too much for an error. Setting the parameter too low has the obvious downside of the LBM not learning anything of value from an example. We chose a relatively standard learning rate of 0.1 for visible-to-hidden edges and 0.0001 for hidden-to-hidden edges.

\begin{figure}[h!]
\center
\includegraphics[scale=.45]{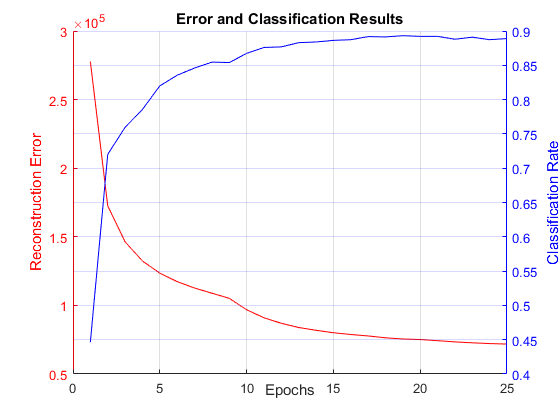}
\caption{A record of reconstruction error and classification rate versus training epochs. Reconstruction error steadily decreases as the classification rate rises, showing that the LBM is learning from the MNIST data.}
\label{fig:errorAndClassRates}
\end{figure}

We wanted to determine if any performance advantage would be gained from using this LBM topology instead of the traditional RBM topology. First we ran a small experiment to confirm that the training of the LBM would behave correctly. Figure \ref{fig:errorAndClassRates} shows the input reconstruction error and the classification rate for a LBM, confirming that it learns the MNIST data. In \ref{table:mnistComparison} we also include a table comparing RBM performance against LBM performance on this MNIST digit classification task. 

The RBM and LBM were both implemented on D-Wave and on MNIST images using the same number of hidden and visible units over ten training epochs. The RBM configuration, as discussed, has no intra-layer connections, whereas the LBM configuration has limited connections between the hidden nodes.

\begin{figure}[h!]
\centering
\begin{tabular}{| c | c | c | c | c |}
\hline
Epoch & RBM Accuracy & LBM Accuracy & RBM Error & LBM Error \\
\hline
1 & 0.4530 & 0.4566 & 275881 & 275102 \\
2 & 0.7277 & 0.7323 & 170908 & 170310 \\
3 & 0.7730 & 0.7913 & 150911 & 146101 \\
4 & 0.8067 & 0.8156 & 139618 & 132333 \\
5 & 0.8162  & 0.8348 & 132283 & 123162 \\
6 & 0.8343 & 0.8468 & 125071 & 116306 \\
7 & 0.8582 & 0.8548 & 114485 & 110730 \\
8 & 0.8657 & 0.8558 & 109022 & 106151 \\
9 & 0.8687 & 0.8553 & 105516 & 101437 \\
10 & 0.8693 & 0.8675 & 102641 & 93042 \\
11 & 0.8705 & 0.8723 & 100452 & 87908 \\
12 & 0.8787 & 0.8783 & 98789 & 84364 \\
13 & 0.8778 & 0.8780 & 97368 & 82016 \\
14 & 0.8747 & 0.8755 & 95365 & 80416 \\
15 & 0.8755 & 0.8771 & 94178 & 79220 \\
16 & 0.8762 &0.8785 & 93237 & 77865 \\
17 & 0.8748 & 0.8781 & 92147 & 76708 \\
18 & 0.8745 & 0.8810 & 91196 & 75916 \\
19 & 0.8743 & 0.8806 & 90312 & 75275 \\
20 & 0.8795 & 0.8831 & 89705 & 74367 \\
21 & 0.8823 & 0.8820 & 89213 & 73643 \\
22 & 0.8780 & 0.8833 & 88426 & 72997 \\
23 & 0.8720 & 0.8838 & 87679 & 72396 \\
24 & 0.8778 & 0.8850 & 87346 & 71952 \\
25 & 0.8755 & 0.8853 & 86850 &  71168 \\
\hline
\end{tabular}
\caption{RBM and LBM performance on the MNIST digit classification task. The LBM tends to label the digits slightly better, and it produces significantly lower reconstruction error than the RBM.}
\label{table:mnistComparison}
\end{figure}

\begin{figure}[h!]
\center
\includegraphics[scale=.4]{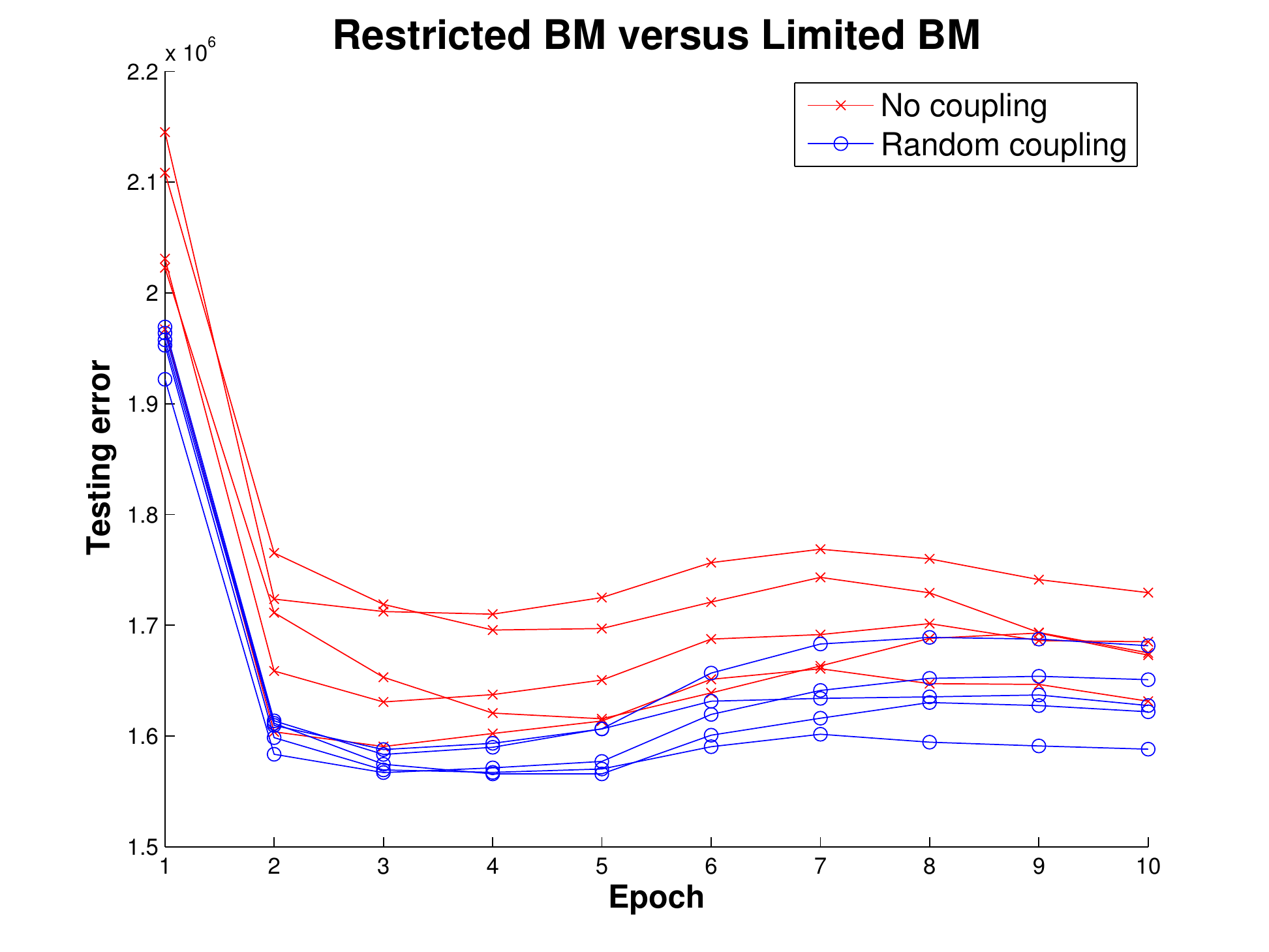}
\caption{Reconstruction error (not normalized) of BMs trained using no intra-layer connections (red) and using random intra-layer connections (blue).}
\label{randomCoupling}
\end{figure}

We initially found that the LBM configuration performed worse than the RBM configuration when we included couplings between nodes in the hidden layer. This was not what we expected, so we introduced a hybrid learning scheme where these intra-layer couplings were redrawn from a random normal distribution for the first three training epochs. From epoch 4 on, the weights were allowed to follow the typical learning rule used in BMs. The results can be seen in the blue series in Figure \ref{randomCoupling}. The choice of using a three epoch duration for randomization was rather arbitrary and the full effects of choosing different durations can be explored in future work. We were primarily interested in providing some randomization while retaining a modest amount of learning time (seven epochs).
% * <tjtt1987@gmail.com> 2017-01-30T18:39:13.797Z:
% 
% Jeremy, can you compare the results of the LBM to RBM. A table of the data would be helpful, as would a description of the results from each topology
% 
% ^.
The final classification rate for one of our trained LBMs was 88.53 percent. Reconstruction error dropped in a regular, expected manner as seen in Figure \ref{randomCoupling}.
In practical terms, we want to compare the cost of training LBMs on quantum devices versus training LBMs on traditional architectures. 

\subsection{HPC}

\begin{figure*} 
\centering
\includegraphics[width=\textwidth]{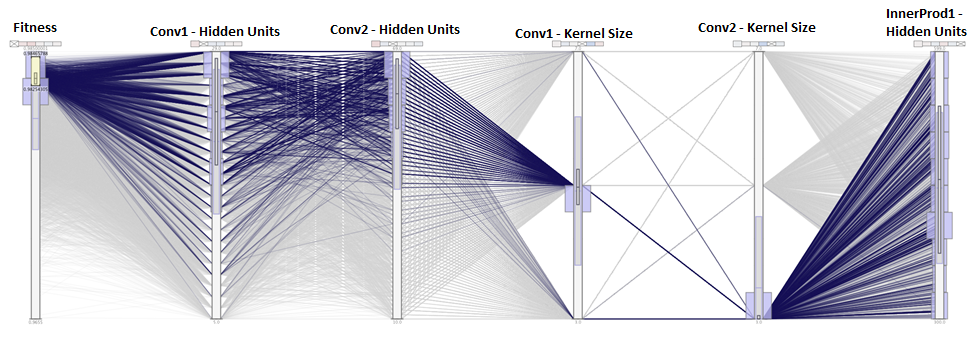}
\caption{Parallel coordinates plot of networks evaluated by the evolutionary algorithm. The best performing networks are highlighted in dark purple such that the best hyper-parameters can be observed. This demonstrates that the best networks utilized smaller kernel sizes than those typically used and that the number of hidden units in the convolutional layers was much more closely tied to performance than the number of hidden units in the inner product layer.}
\label{fig:learned_hyperparameters}
\end{figure*}

% Results of using Menddl to find an MNST topology

\begin{figure}[h!]
\center
\includegraphics[width=0.5\columnwidth]{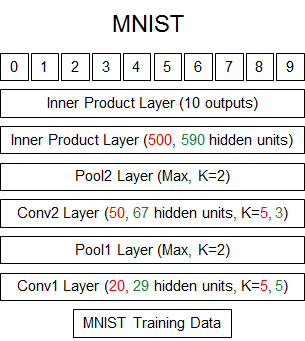}
\caption{Network being optimized. This figure shows the typical hyper-parameters used (red) and the hyper-parameters learned through the evolutionary algorithm (green).}
\label{fig:standard_vs_optimized}
\end{figure}

We used the Titan computer and the MENNDL system to discover a near optimal topology of a deep network trained on the MNIST handwritten digit dataset \cite{mnist} by utilizing the method presented in \cite{young2015}. 
The hyper-parameters optimized were the kernel size, the number of hidden units for each of the convolutional layers, and the number of hidden units in the fully connected layer. The structure of the network is shown in Figure \ref{fig:standard_vs_optimized}.
Utilizing $500$ nodes of Titan, the evolutionary algorithm was trained for $32$ generations with $500$ individuals in the population allowing us to evaluate $16,000$ networks.  Each hyper-parameter is encoded as an integer gene, and the range of this integer is limited in order to avoid evaluating hyper-parameter values that are not of interest.
A single node of Titan evaluates the core of the evolutionary algorithm and distributes the fitness function to the rest of the nodes to evaluate the network using Titan's GPUs.

The optimal network, shown in Figure \ref{fig:standard_vs_optimized}, demonstrates some significant differences from the starting network. The optimal network achieved 98.5 percent classification accuracy, representing a slight increase over the baseline network. This demonstrates that accuracy can be improved by optimizing hyper-parameters even on networks and datasets that have been widely used.  Figure \ref{fig:learned_hyperparameters} highlights the best performing networks and shows their corresponding hyper-parameters. 
It is interesting to note that the best performing networks had a wide variety of values for the number of hidden units in the fully connected layer. 
However, there was little variation in the kernel size of the convolutional layers which indicates that the performance of the network is much more sensitive to this parameter, and the kernel size of the second layer converged to a much smaller value than the value that is typically used. 
This indicates that even for very well studied problems, i.e., MNIST, the networks typically used are not optimal since it is difficult to find the correct hyper-parameters without an automated search process and significant computing resources.

\subsection{Neuromorphic}
% Results of using SSN and FPGA on MNST
% should we include MNIST accuracy results for NIDA? Referencing prior pubs of course?
%  the memristor based simulation is for a network that was converted from NIDA so I think this would be OK -- GSR

% here are some results for memristive NIDA -- GSR (will add some background on memristive neuromorphic)

This experiment was done in two parts, the first was to implement a SNN using NIDA to demonstrate a neuromorphic solution to the MNIST problem is feasible. The second part is to simulation the characteristics of this SNN implemented on memristive hardware. 

\begin{figure} [h!]
  \centering
  \includegraphics[width=.7\textwidth]{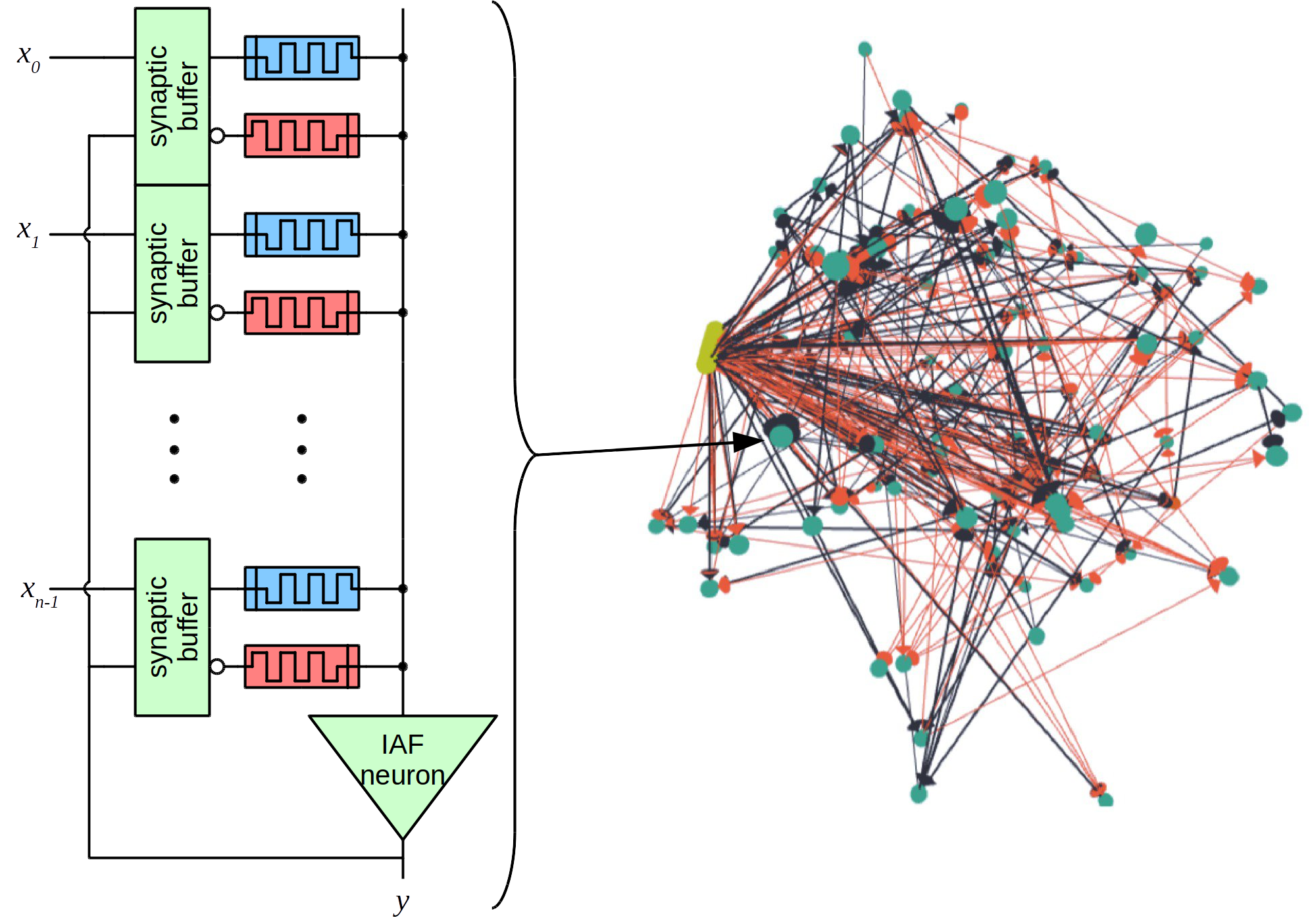}
  \caption{An example memristive neuromorphic circuit with integrated-and-fire (IAF) neuron (left) and a NIDA network trained to classify digit `0' MNIST images (right).}
  \label{fig:memrNIDA}

\end{figure}

We started by simulating a memristive implementation of a NIDA network trained to classify MNIST images (Figure \ref{fig:memrNIDA}). The NIDA network itself was generated using evolutionary optimization and was part of an ensemble of networks that classifies MNIST images with an accuracy of approximately 90 percent. Energy consumption was also estimated for a NIDA representation of this network where synaptic elements are implemented using metal-oxide memristors.
 
The memristive device technology assumed for this simulation is characterized by a low resistance state (LRS) of 60k$\Omega$, about an order of magnitude larger than the resistance of a typical deep-submicron CMOS transistor. This relatively high LRS for the memristor is desirable such that the CMOS channel resistance can effectively be neglected. The on-off ratio is assumed to be 10, providing a high resistance state (HRS) of 600k$\Omega$. Such characteristics for LRS, HRS and the associated on-off ratio have been observed for a range of memristive devices, including hafnium-oxide (HfO$_2$)\cite{Cady16}, tantalum-oxide (TaO$_2$), and titanium-oxide (TiO$_2$), to name a few. All of these memristive material stacks consist of an oxide layer sandwiched between two metallic layers. Depending on the polarity and magnitude of an applied voltage bias, the oxide layer transitions between being less or more conductive, providing the switching characteristics desirable for representing synaptic weights.
% Gangotree: please add 1-2 more refs justifying why we assume the above. 
% * <tjtt1987@gmail.com> 2017-01-30T20:44:34.206Z:
% 
% > The memristive device technology assumed for this simulation is characterized by a 60k$\Omega$ low resistance state and an on-off ratio of just 10. 
% 
% Garrett, can you explain this a bit more. How does the a 60kOhm resistance compare with normal CMOS resistance?
% Also what is an on-off ratio?
% 
% ^ <tjtt1987@gmail.com> 2017-02-24T16:10:28.338Z.

%While the on-off ratio is fairly low, this model is representative of characteristics achievable for experimentally observed hafnium-oxide memristors \cite{Cady16}. 

% * <tjtt1987@gmail.com> 2017-01-31T15:36:45.676Z:
% 
% Garrett, can you write a few sentences about hafnium-oxide for a non-EE person?
% 
% 
% ^ <tjtt1987@gmail.com> 2017-02-24T16:10:31.611Z.
 
Our memristive NIDA simulation setup also includes analog integrate-and-fire neurons, implemented using a 65nm CMOS process technology. Neuromorphic elements (neurons and synapses) were simulated using Cadence Spectre and system-level energy and power estimates were calculated using a high-level simulator written in C++.

The memristive NIDA simulation is based on two significant steps. Initially the Evolutionary Optimization based simulator is used to generate optimized networks for the low level simulation. Then the transistor level simulation is done using Cadence Spectre simulator. Some power estimates are collected for the design components in different conditions. Those are used to calculate the total energy consumed during the simulation of the application.

To collect energy data, we considered three different phases for neuron and those are fire, accumulation, and idle phase. Using Cadence simulator energy of each phase has been determined as per spike energy. Similar method is followed for synapses as well. But here we considered the active and passive phases of a synapse. And for the programmable delay chain, we collected energy data for each spike existing in the delay chain. The EO based memristive NIDA simulator is used to generate network with some activity factors for specific applications. For instance, MNIST dataset was used to verify in recognizing the handwritten characters and the total energy of the system was calculated. To calculate the average energy for a single image run, the total energy consumed by the network is calculated for the total number of runs (which, for example, is 10000 for MNIST dataset application). The network has 128 neurons and 357 synapses associated with 357 programmable delays. The neurons and synapses are mostly analog circuit components but the programmable logic delay consists of digital components. That is why the energy consumed is mostly because of the digital parts. The average power observed for the total network with 16.67 MHz clock speed is 304.3mW and the respective energy is 18.26nJ (including the digital programmable delays). But if we consider the core analog neuromorphic logic the energy per spike is 5.24nJ and the average power is 87.43mW which is consistent with similar memristor-based neuromorphic systems \cite{liu2016memristor}. Research has also shown that memristive neuromorphic systems are typically $20\times$ more energy-efficient than their CMOS counterparts \cite{Liu2016}, and our results are consistent with this estimation. Further improvements in energy-efficiency are possible through the use of memristors with a higher on-off ratio and/or higher low resistance  state. Ultra low-power CMOS circuit design techniques (i.e., sub-threshold operation) can also be used to further reduce the power consumption of CMOS neurons. Thus, a CMOS-memristive neuromorphic implementation is particularly well suited for energy-constrained, resource limited application domains.

%A memristive NIDA network used to classify a particular MNIST image (the digit `0') was found to consume an average power of 1.72mW. For the full 500 cycles required to classify the MNIST image, including loading the image and allowing the network to process the data, our memristive implementation was found to consume a total energy of 710nJ. These results are consistent with results determined for similar memristor-based neuromorphic systems \cite{Taha15}. Research has also shown that memristive neuromorphic systems are typically $20\times$ more energy-efficient than their CMOS counterparts \cite{Liu2016}, and our results are consistent with this estimation. Further improvements in energy-efficiency are possible through the use of memristors with a higher on-off ratio and/or higher low resistance  state. Ultra low-power CMOS circuit design techniques (i.e., sub-threshold operation) can also be used to further reduce the power consumption of CMOS neurons. Thus, a CMOS-memristive neuromorphic implementation is particularly well suited for energy-constrained, resource limited application domains.
% * <tjtt1987@gmail.com> 2017-01-31T15:39:00.128Z:
% 
% > For the full 500 cycles required to classify the MNIST image, including loading the image and allowing the network to process the data, our memristive implementation was found to consume a total energy of 710nJ.
% 
% Garrett, can you explain these steps in greater deal? Enough so that someone can replicate what you have done.
% 
% ^ <tjtt1987@gmail.com> 2017-02-24T16:13:17.390Z.

\section{Discussion}

The goal of this study is to explore ways of addressing some of the current limitations of deep learning, namely, 
complex topologies that are closer representations of nature, automatically configuring a the hyperparameters of a network, and natively implement deep learning model using neuron and synapse hardware.

We use three architectures, quantum computing, high performance computer, and neuromorphic computing and three different deep learning models, LBM, CNN, and SNN to 
address these issues.

The quantum approach allows the deep learning network topologies to be much more complex than is feasible with conventional computers. The results show training convergence with a high number of intra-layer connections, thus opening the possibility of using much more complex topologies that can be trained on a quantum computer. There is not a time-based performance penalty for increased intra-layer connections, although, there may be the need to do more sampling in order to reduce potential errors. 

HPC's contribution to the problem focuses on automatically developing an optimal network topology to create a high performing network. Many of the topologies used today are developed through trial and error methods. This approach works well with standard research datasets since the research community can learn and publish the topologies that produce the highest accuracy networks for these data. It is a different matter when working with datasets that have not been widely studied. The HPC approach provides a way to optimize the hyper-parameters of a CNN, saving significant amounts of time when working on new datasets. 

The neuromorphic contribution to this problem is to provide a native implementation and a low-power memristor-based hardware to implement of a SNN. The network has the potential to have broader connections than a CNN and the ability to dynamically reconfigure itself over time. There are many benefits to neuromorphic computers (including robustness, low energy usage, and small device footprint) that could be useful in a real-world environment today if we had a mechanism for finding good network solutions to deploy on those devices.  

Reviewing the results of the three experiments opens the possibility of using these three architectures in tandem to create powerful deep learning systems that are beyond our current capabilities. Practically, the current quantum computer is quite limited in the size and scope of the problem it can address, but the ability to train a very complex deep learning network gives it a very interesting potential. 
It could be used to generate weights for very complex networks that are untrainable using current systems opening the possibilities of potentially solving more complex and challenging problems. However, the scalability of a quantum machine is a real concern. As we observed, limiting the size of the input layer to 1000 qubit severely limits the size of a problem that can be analyzed using this approach. We believe that the best use for complex networks may be as higher layers in a CNN. These layers usually are combining fairly rich features, and may benefit from increased inter-layer connections. These layers usually have a smaller input size than the original input, which eases the scalability concerns of this approach, and may improve over all accuracy.

The HPC approach of automatically finding optimal deep learning topologies is a fairly robust and scalable capability, though quite expensive in development and computer costs. Having the ability to use deep learning methods on unstudied datasets (experimental scientific data) can provide a huge time savings and analytical benefit to the scientific community.  

The neuromorphic approach is limited by the lack of robust neuromorphic hardware and algorithms, yet it holds the potential of analyzing complex data using temporal analysis and very low power hardware. One of the most compelling aspects of this approach is the combination of a SNN and neuromorphic hardware that can analyze the temporal aspects of data. The MNIST problem does not have a temporal component, but one can imagine a dataset that has both image and temporal aspects, such as a video. A CNN approach has been shown to perform well on the image side, perhaps a SNN can provide increased accuracy by analyzing the temporal aspects as well. 

\begin{figure*} 
\centering
\includegraphics[width=0.7\columnwidth]{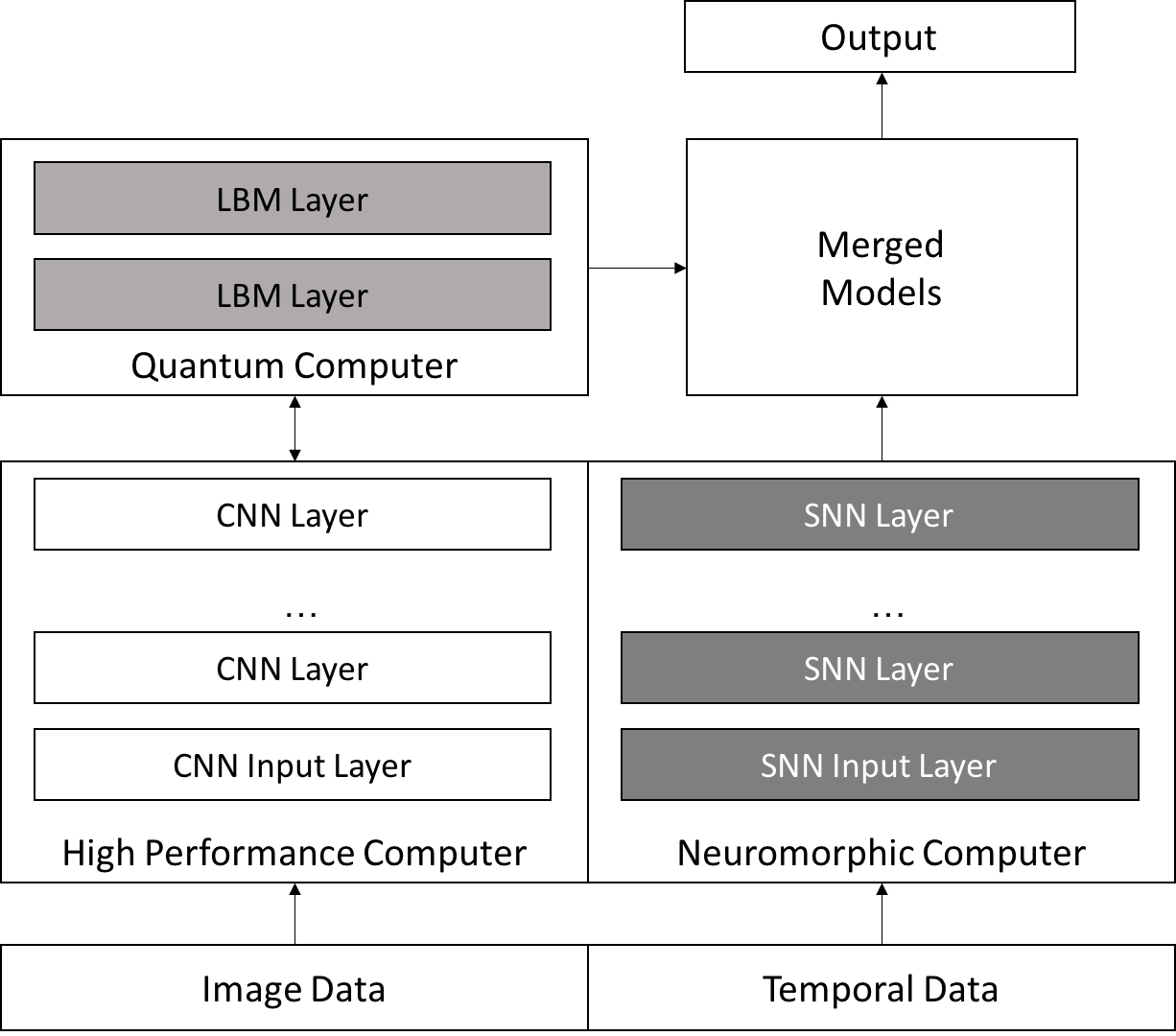}
\caption{A proposed architecture that shows how the three approaches, quantum, HPC, and neuromorphic can be used to improve a deep learning approach}
\label{fig:Proposed_Architecture}
\end{figure*}

These experiments provide  valuable insights into deep learning by exploring the combination of three novel approaches to challenging deep learning problems. We believe that these three architectures can be combined to gain greater accuracy, flexibility, and insight into a deep learning approach. Figure \ref{fig:Proposed_Architecture} shows a possible configuration of the three approaches that addresses the three deep learning challenges we discussed above. The high performance computer is used to create a well performing CNN on image type data. The final layer or two is then processed by the quantum computer using a LBM network that contains greater complexity than and CNN. The temporal aspects of the data are modeled using a SNN, and the ensemble models are then merged and an output produced. Our belief is that this approach has the potential to yield greater accuracy than existing CNN models.

\subsection{Future Work}
Our next step is to test the hypothesis that this proposed architecture does indeed provide greater accuracy, flexibility, and insight into a dataset than can be derived from a traditional CNN approach. We will apply this proposed architecture to a large scientific dataset, and compare the results of a traditional approach to this proposed architecture.

\section{Conclusion}

Current Deep Learning networks are loosely based on this neural model of human perception and have been highly optimized using CNNs trained on large clusters of GPUs. This technology has been instrumental in solving problems that have challenged researchers for years, such as object and facial recognition within photographs. The topology of these CNN networks consist of convolutional layers with shared weights and fully connected layers, without inter-layer connections, which while powerful, are quite simplistic.

This paper addresses three main limitations in deep learning: 1) training models with complex topologies that contain intra-layer connections; 2) automatically determining an optimal configuration for a network topology; and 3) implementing a complex topology in native hardware. To address these problems we explored a simple deep learning problem on three different architectures: quantum, high performance, and neuromorphic computers. These architectures address the three problems: complex topologies with quantum computing; network topology optimization with high performance computing; and low-power implementation with neuromorphic computing. 

Given input size limitations of 1,000 qubits, we use the MNIST dataset for this evaluation, and use neuron models and topologies that are best suited to the architectures: CNN for HPC; SNN for neuromorphic; and BMs for quantum. 

Our results from these three experiments demonstrate the possibility of using these three architectures to solve complex deep learning networks that are currently untrainable using a von Neumann architecture. 

The quantum computer experiment demonstrated that a complex neural network, i.e., one with intra-layer connections, can be successfully trained on the MNIST problem. This is a key advantage for a quantum approach and opens the possibility of training very complex networks. A high performance computer can be used to take the complex networks as building blocks and compare thousands of models to find the best performing networks for a given problem. And finally, the best performing neural network and weights can be implemented into a complex network of memristors producing a low-power hardware device. This is a capability that is not feasible with a von Neumann architecture. This holds the potential to solve much more complicated problems than can currently be solved with deep learning.

We propose a new deep learning architecture based on the unique capabilities of the quantum, high performance, and neuromorphic approaches presented in this paper. This new architecture addresses the three main limitations we see in current deep learning methods, and holds the promise of higher classification accuracy, faster network creation times, and low power, native implementation in hardware.

%\end{document}  % This is where a 'short' article might terminate

%ACKNOWLEDGMENTS are optional
\section{Acknowledgments}
This material is based upon work supported by the U.S. Department of Energy, Office of Science, Office of Advanced Scientific Computing Research, Robinson Pino, program manager, under contract number DE-AC05-00OR22725.
This research used resources of the Oak Ridge Leadership Computing Facility, which is a DOE Office of Science User Facility supported under Contract DE-AC05-00OR22725.
\bibliographystyle{abbrv}
\bibliography{references}  % sigproc.bib is the name of the Bibliography in this case
% ACM needs 'a single self-contained file'!
%\balancecolumns
\
%\balancecolumns % GM June 2007
% That's all folks!
\end{document}